\documentclass[runningheads]{llncs}
\usepackage[T1]{fontenc}
\usepackage{graphicx}
\usepackage{cite}          
\usepackage{hyperref}      
\usepackage{url}           

\usepackage{amsmath, amssymb, amsfonts}  
\usepackage{mathtools}      
\usepackage{commath, bm}    
\usepackage{dsfont}         

\usepackage{algorithm}       
\usepackage{algpseudocode}   

\usepackage{float}
\usepackage{placeins}
\usepackage{graphicx}         
\graphicspath{{figures/}}     
\usepackage{caption}          
\usepackage{subcaption}       
\usepackage{wrapfig}          
\usepackage{booktabs}         
\usepackage{multirow}         
\usepackage{ctable}           
\usepackage{tablefootnote}    
\usepackage{tabularray}       
\usepackage{xcolor, colortbl} 
\usepackage{float}            

\usepackage{enumitem}         
\usepackage{comment}          
\usepackage{xspace}           

\newcommand{\Krum}{\texttt{Krum}\xspace}
\newcommand{\mKrum}{$\texttt{mKrum}$\xspace}
\newcommand{\rKrum}{\texttt{rKrum}\xspace}
\newcommand{\ArKrum}{$\texttt{ArKrum}$\xspace}
\newcommand{\Mean}{\texttt{Mean}\xspace}
\newcommand{\kscore}{\texttt{Score\textsubscript{Krum}}\xspace}

\newcommand{\Fig}{Fig.\hspace{-0.15em}}
\newcommand{\Alg}{Alg.\hspace{-0.15em}}
\newcommand{\lspace}{~\hspace{-0.20em}}

\begin{document}
\title{Secure and Private Federated Learning: Achieving Adversarial Resilience through Robust Aggregation\thanks{This work used resources of the O’Donnell Data Science and Research Computing Institute at Southern Methodist University.}}
\titlerunning{}
%
\author{Kun Yang\orcidID{0009-0007-2145-596X} \and
Neena Imam\orcidID{0000-0001-8860-4738}}
\authorrunning{K. Yang and N. Imam}
%
\institute{
Southern Methodist University, Dallas, TX 75205, USA\\ 
\email{\{kunyang,nimam\}@smu.edu}
}

\maketitle              
%


\begin{abstract}
Federated Learning (FL) enables collaborative machine learning across decentralized data sources without sharing raw data. It offers a promising approach to privacy-preserving AI. However, FL remains vulnerable to adversarial threats from malicious participants, referred to as Byzantine clients, who can send misleading updates to corrupt the global model. Traditional aggregation methods, such as simple averaging, are not robust to such attacks. More resilient approaches, like the \Krum algorithm, require prior knowledge of the number of malicious clients, which is often unavailable in real-world scenarios. To address these limitations, we propose Average-\rKrum (\ArKrum), a novel aggregation strategy designed to enhance both the resilience and privacy guarantees of FL systems. Building on our previous work (\rKrum), \ArKrum introduces two key innovations. First, it includes a median-based filtering mechanism that removes extreme outliers before estimating the number of adversarial clients. Second, it applies a multi-update averaging scheme to improve stability and performance, particularly when client data distributions are not identical. We evaluate \ArKrum on benchmark image and text datasets under three widely studied Byzantine attack types. Results show that \ArKrum consistently achieves high accuracy and stability. It performs as well as or better than other robust aggregation methods. These findings demonstrate that \ArKrum is an effective and practical solution for secure FL systems in adversarial environments.

\end{abstract}

\section{Introduction}
\begin{figure}[!ht]
    \centering
    \includegraphics[width=0.99\linewidth, trim=1.1cm 1.8cm 3.8cm 2.0cm, clip]{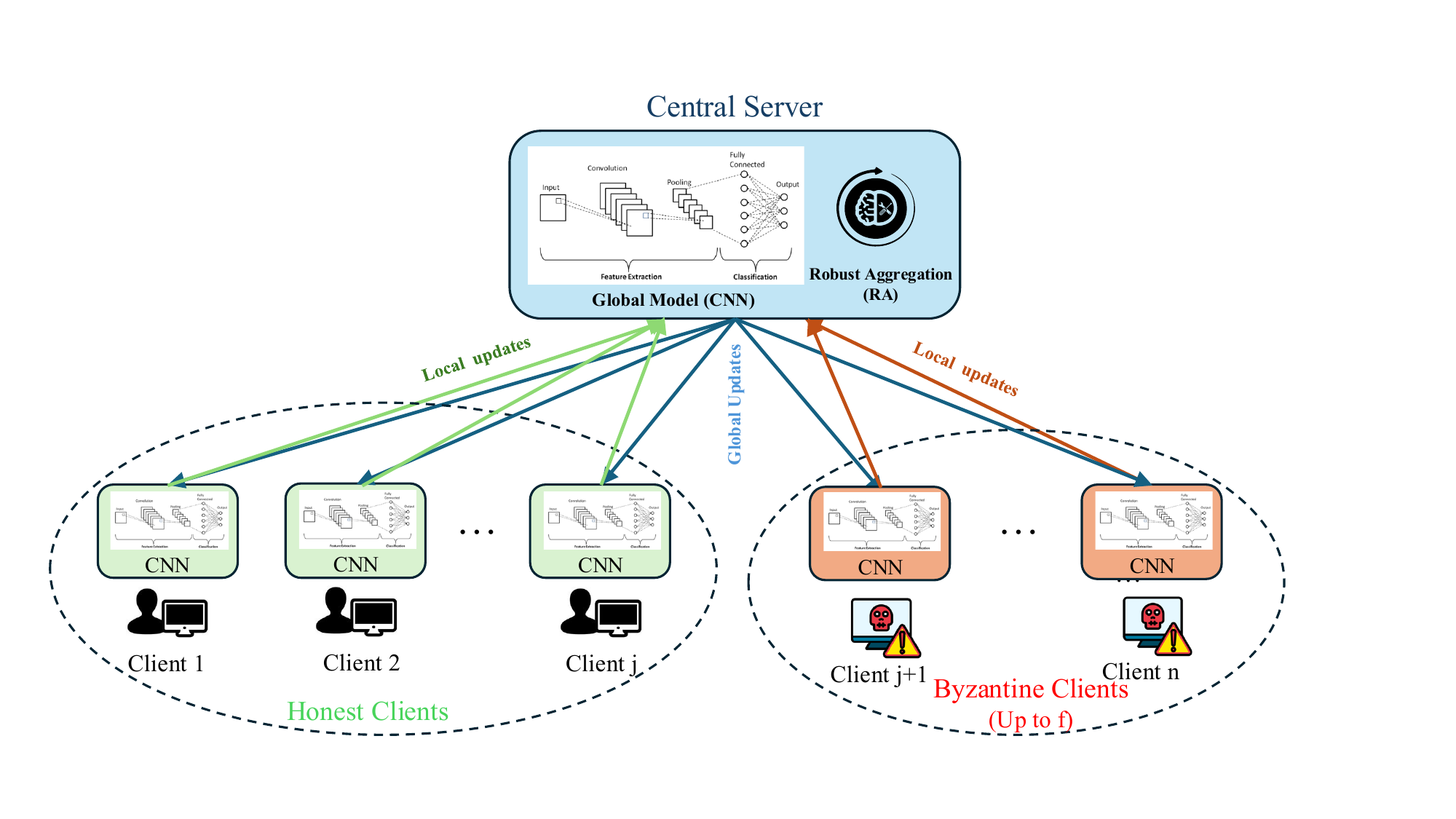}
    \caption{Federated Learning Architecture.}
    \label{fig:fl_architecture}
    \vspace{-0.8cm}
\end{figure}

The computing ecosystem is evolving toward a distributed paradigm, driven by the exponential growth of data, the need for real-time processing, and the proliferation of edge devices. This trend decentralizes computation across cloud, edge, and device layers, transforming how computational workloads are handled. Machine Learning (ML) must adapt to this distributed environment, supporting real-time, in situ data analysis at the edge. A promising approach to this heterogeneous environment is Federated Learning (FL) which enables collaborative ML across decentralized data sources without sharing raw data, thus ensuring data privacy. FL has attracted increasing attention from both academia and industry due to this privacy-preserving capabilities~\cite{mcmahan2017communication}. A typical FL system consists of two main components: distributed clients and a central server, as shown in \Fig~\ref{fig:fl_architecture}. In this context, clients generally refer to processing nodes that locally train models on private data. The overall process works as follows. First, each client trains a local model on its local data and sends the resulting model updates to a central server. The server then aggregates these model updates using an aggregation method to update a global model. This process is repeated iteratively until a predefined convergence criterion is met. In the classical FL, the aggregation method simply averages the model updates for each coordinate, known as federated averaging or coordinate-wise Mean (\Mean)~\cite{mcmahan2017communication}. 
\vspace{-0.6cm}
FL is vulnerable to malicious attacks because it relies on decentralized contributions, making it susceptible to poisoned or manipulated model updates from compromised participants ~\cite{karimireddy2021learning}. The malicious participants in FL are referred to as Byzantine clients. The Byzantine clients, compromised and controlled by attackers, attempt to disrupt the FL-based applications by sending arbitrary or malicious model updates~\cite{moshawrab2023reviewing}.  

To address this security challenge, instead of using the standard aggregation method in the classical FL frameworks, \textit{robust} aggregation algorithms have been proposed \cite{moshawrab2023reviewing}. 
A robust aggregation method in FL is a technique where the server aims to effectively aggregate the model updates received from distributed clients while minimizing the impact of Byzantine clients.
The \Krum algorithm was proposed by Blanchard et al. as a robust alternative to the classical \Mean aggregation method \lspace\cite{blanchard2017machine}, and it has since become a widely adopted.
Specifically, the \Krum algorithm remains effective in FL with \( n \) clients, where up to \( f \) clients may be Byzantine, under the constraint \!\(2\!+\!2f\!<\!n\). 
The standard \Krum algorithm works as follows:
\begin{itemize}[topsep=3pt, itemsep=0pt, parsep=0pt, partopsep=0pt]
    \item \textbf{Step~1:} Compute the squared Euclidean distance \( d_{ij}\! =\! \| \mathbf{u}_i\!-\!\mathbf{u}_j \|_2^2 \) between each pair of client model updates, where \( \mathbf{u}_i \) and \( \mathbf{u}_j \) denote the model updates from clients \( C_i \) and \( C_j \), respectively.
    
    \item \textbf{Step~2:} For each client \( C_i \), sort the distances \( d_{ij} \) to all other \( n\!-\!1 \) clients in ascending order. Then compute the sum of the first \(n\!-\!f\!-2\) distances.  This sum is referred to as the \Krum score (\kscore) for the client: 
    \(
    \kscore(\mathbf{u}_i)\!=\! \sum_{j \in \mathcal{N}_i} d_{ij}, 
    \)
    where \( \mathcal{N}_i \) denote the indices of the \( n\! -\! f\! -\! 2 \) clients nearest to  \( C_i \). 
    
    \item \textbf{Step~3:} Select the model update \( \mathbf{u}^{*} \) with the smallest \kscore as the final aggregated result:  
    \(
    \mathbf{u}^*\! =\! \arg\min_{\mathbf{u}_i \in \{\mathbf{u}_1, \ldots, \mathbf{u}_n\}} \kscore(\mathbf{u}_i).
    \)
\end{itemize}

Based on the main steps of the \Krum algorithm, we observe that Step 2 requires prior knowledge of the number of Byzantine clients \(f \). However, in practice, this parameter is challenging to determine in advance due to the complex and dynamic nature of distributed systems.
In our prior work, we proposed a refined \Krum algorithm called \rKrum that eliminates the need for hard-coded threat models \cite{yang2025resilient}. Our \rKrum algorithm formulated the estimation of \( f \) as a change point detection problem and employed a Sum-of-Squared-Errors (SSE)-based segmentation method to automatically estimate \( f \) without introducing any additional parameters. Despite its advantages, we observed that \rKrum tends to underestimate \( f \), especially when Byzantine clients generate extreme model updates. Consequently, some malicious model updates may be mistakenly included when computing the \kscore. Moreover, similar to the \Krum algorithm, our \rKrum algorithm selected the model update with the lowest \Krum score, discarding all others (even updates generated by honest clients). This strategy could cause instability and slow convergence, especially in Non-Independent and Identically Distributed (Non-IID) settings, where client data distribution can differ significantly. 
To overcome these limitations, we propose an enhanced variant called Average-\rKrum (\ArKrum). 
Our new contributions are as follows:
\begin{itemize}[topsep=3pt, itemsep=0pt, parsep=0pt, partopsep=0pt]
    \item \textbf{Extreme Value Filtering}: To mitigate the impact of the extreme updates on our SSE-based estimation method, we introduce a median-based filtering method that effectively removes extreme values before the estimation.
    
    \item \textbf{Stability Enhancement via Multi-Aggregation}: To further improve the stability of our \rKrum algorithm, we adopt a strategy inspired by multi-\Krum (\mKrum)~\cite{blanchard2017machine}, where multiple client updates are averaged instead of selecting a single update with the lowest \kscore, as in the standard \Krum algorithm. 
\end{itemize}
We evaluate the effectiveness of our proposed \ArKrum algorithm on two public datasets under three widely studied Byzantine attacks. Experimental results demonstrate that \ArKrum achieves comparable global model accuracy to that of both \Krum and \mKrum in most cases.
As anticipated, \ArKrum and \mKrum are more stable than \rKrum and \Krum, which aggregate by selecting only a single update.
    
The remainder of the paper is organized as follows. In Section 2, we review related work on various aggregation algorithms. In Section 3, we introduce our proposed algorithm, \ArKrum.  In Section 4, we provide a comprehensive numerical analysis. We conclude the paper in Section 5 with directions for future work.

\section{Related Work}
Since Blanchard et al. introduced the \Krum algorithm as a robust alternative aggregation method, it has inspired many research studies in this direction \cite{blanchard2017machine}. For instance, Wang et al. proposed a blockchain-based FL framework to enhance the robustness of FL in smart transportation systems \cite{wang2024blockchain}. They integrated \Krum with homomorphic encryption, enabling ciphertext-level model aggregation and filtering. Their results demonstrated that the proposed scheme satisfied privacy and security requirements while improving model performance. Similarly, Garcia et al. explored the integration of blockchain and \Krum to enhance the resilience of FL systems against Byzantine attacks \cite{garcia2025krum}. 
Another important application area is the Internet of Medical Things (IoMT), which also faces security and privacy challenges. 
For instance, Fahim et al. proposed a resource-efficient FL framework that integrated \Krum for intrusion detection in IoMT environments \cite{fahim2025resource}. Additionally, Colosimo et al. aimed to improve the standard \Krum algorithm by integrating it with coordinate-wise median \cite{colosimo2023median}. In Colosimo's integrated framework, each client first computed its \kscore, and then the server selected a subset of clients with the lowest scores to perform coordinate-wise median aggregation. While their experimental results demonstrated that the proposed method can improve robustness of the standard \Krum algorithm, it still required prior knowledge of the number of Byzantine clients. Despite its effectiveness against Byzantine attacks, the standard \Krum algorithm still has its limitations, e.g., requiring the number of Byzantine clients \( f \) to be specified in advance. In our prior work, we proposed a refined \Krum algorithm called \rKrum, which automatically estimates \( f \) without introducing any additional parameters \cite{yang2025resilient}. Despite its advantages, we observed that our \rKrum algorithm tends to underestimate \( f \) and becomes unstable in Non-IID settings. To address these limitations, we introduce an enhanced version of \rKrum called Average-\rKrum (\ArKrum) in this paper.

\section{Proposed Algorithm Average-\rKrum (\ArKrum)}
In this section we describe in detail the improvements made to our prior \rKrum algorithm for better stability. We implement a median-based filtering algorithm to remove extreme model updates to mitigate the underestimation of \(f\). We further improve the stability of \rKrum by aggregating multiple client updates (inspired by \mKrum). By combining these two enhancements, we develop an improved version of \rKrum, which we call Average-\rKrum (\ArKrum).

\subsection{Federated Learning Framework}
Our FL environment has \( n \) clients, where up to \( f \) of these clients are Byzantine clients, under the constraint \(2\!+\!2f\!<\!n \), as illustrated in \Fig~\ref{fig:fl_architecture}. The honest clients are shown in green, while the Byzantine clients are shown in orange. The goal is to learn a global model in the presence of Byzantine clients. Example models may be Convolutional Neural Networks (CNNs) for image data or Fully-connected Neural Networks (FNNs) for tabular data.
Each honest client \( C_i \) independently trains its local model on its local data and transmits the model updates \( \mathbf{u}_i \in \mathcal{R}^{d} \) to a central server, ensuring that no raw data are shared between clients. Here, \( d \) is the dimensionality of the model update. The Byzantine clients aim to disrupt the global model's learning process, either by injecting random noise or introducing misleading information before sending the modified updates to the server. 
To address the challenge posed by the Byzantine clients, the server employs robust aggregation methods for model updates. 
\vspace{-0.5cm}
\begin{algorithm}[H]
\caption{Filter Extreme Model Updates with Median}
\label{alg:extreme_value_filtering}
\begin{algorithmic}[1]
\Require Sorted array of squared Euclidean distances \( D_i^{\prime} \gets [d'_{i1}, d'_{i2}, \ldots, d'_{in}] \)
\Ensure Filtered array \( D_i^{*} \)
\Procedure{Filter\_Extreme\_Values}{$D_i$}
    \State \( \text{mid} \gets \left\lfloor \frac{n}{2} \right\rfloor \), \( \text{median} \gets d'_{i\,\text{mid}} \), \( \Delta_{\max} \gets \text{median} - d'_{i1} \)
    \State  \( \tau \gets \text{median} + \Delta_{\max} \), \( j_{\max} \gets n \)
    \State \# Find the first extreme value location
    \For{\( j = \text{mid} + 1 \) \textbf{to} \( n \)}
        \If{\( d'_{ij} > \tau \)}
            \State \( j_{\max} \gets j-1 \)
            \State \textbf{break}
        \EndIf
    \EndFor
    \State \( n^{\prime} \gets j_{\max} \)
    , \( D_i^{*} \mathrel{\gets}  \{d'_{i1}, d'_{i2}, \ldots, d'_{i n^{\prime}}\} \)
    \State \Return \( D_i^{*} \)
\EndProcedure
\end{algorithmic}
\end{algorithm}
\vspace{-0.8cm}
        
\subsection{Filtering Extreme Values}
As discussed in our prior work, \rKrum algorithm formulates the estimation of \( f \) as a change point detection problem. We employ an SSE-based segmentation method to automatically estimate \( f \) without introducing any additional parameters. Specifically, our SSE-based estimation method minimizes the total squared error of the left and right segments of the sorted distances when estimating \( f \)~\cite{yang2025resilient}. We observed that this method can skew the estimated change point towards extreme model updates. We assume no prior knowledge of Byzantine behavior, i.e., we allow these Byzantine clients to generate any arbitrary, even extreme, model updates. Consequently, the SSE-based estimation of \( f \) may be underestimated, i.e., we may include some Byzantine updates when computing the \kscore. This is a problem we aim to avoid in our \ArKrum algorithm introduced in this paper. We propose a median-based filtering algorithm to exclude the extreme updates as detailed in \Alg~\ref{alg:extreme_value_filtering} (Filter Extreme Model Updates with Median). This median approach is applied before the SSE-based segmentation method in our \rKrum algorithm (prior work). Specifically, given a sorted array of distances \( D_i^{\prime} \), we first compute the median at the lower index \( \text{mid} = \lfloor n/2 \rfloor \). Next, we calculate the maximum distance \( \Delta_{\text{max}} \) from the median to the smallest value on the left side (i.e., \( d'_{i1} \), the first value of \( D_i^{\prime} \) since \( D_i^{\prime} \) is sorted). 
Based on this, we define a threshold \( \tau = \Delta_{\text{max}} + d'_{i1} \). We then filter out any distances from \( D_i^{\prime} \) that exceed this threshold. More details are provided in \Alg ~\ref{alg:extreme_value_filtering}. 
The rationale for using the median as the central reference point is that we assume that no more than half of the model updates are generated by Byzantine clients. Therefore, the median update is guaranteed to be generated by one of the honest clients, whom we can trust.
By removing the extreme updates potentially generated by Byzantine clients, we then pass the filtered array \( D_i^{*} \) to our \rKrum algorithm to obtain a more accurate estimate of \( \hat{f}_i \) for each client \( C_i \).

\vspace{-0.2cm}
\begin{algorithm}[H]
\caption{Parameter-free \Krum for Robust Aggregation (\texttt{\ArKrum})}
\label{alg:ArKrum}
\begin{algorithmic}[1]
\Require \( U = \{\mathbf{u}_1, \mathbf{u}_2, \ldots, \mathbf{u}_n\} \) (represented the set of model updates received from \( n \) clients, where each \( \mathbf{u}_i \in \mathbb{R}^d \) represents a model update of dimension \( d \).
\Ensure Aggregated model update \( \overline{\mathbf{u}} \)
\Procedure{rKrum}{$U$}
    \State \# {Step 1: Compute pairwise Euclidean distance}
    \For{each pair of updates \( (\mathbf{u}_i, \mathbf{u}_j) \) where \( i \neq j \)}
        \State \( d_{ij} = \| \mathbf{u}_i\! -\! \mathbf{u}_j \|_2^2 = \sum_{k=1}^{d} ( u_{ik}\! -\! u_{jk} )^2 \)
    \EndFor

    \State \# {Step 2: Estimate \( \hat{f} \) and compute \Krum score}
    \For{each update \( \mathbf{u}_i \)}
        \State    \# Sort pairwise distances \( d_{i\cdot} \) in ascending order
        \State \( D \gets \{ d_{i1}^{\prime}, d_{i2}^{\prime}, \ldots, d_{in}^{\prime} \} \) 
        \State \( D' \gets \textbf{FILTER\_EXTREME\_VALUES}(D) \) \quad    \# \Alg~\ref{alg:extreme_value_filtering}
        \State \( \hat{f}_i \gets \textbf{ESTIMATE\_F}(D') \)  \quad   \# \rKrum in \cite{yang2025resilient}
        \State \# Compute \Krum score:
        \State \(  \kscore_({\mathbf{u_i}}) = \sum_{j \in \mathcal{N}_i} d_{ij}^{\prime}\) \quad \# \( \mathcal{N}_i \) is  the indices of the top \( n\! -\!  \hat{f}_i\! -\! 2 \) distances.
    \EndFor

    \State \# {Step 3: Select the update with the minimal \Krum score}
    \State \( \mathbf{u}_{i^{*}} = \underset{\mathbf{u_i} \in \{\mathbf{u_1}, \ldots, \mathbf{u_n}\}}{\arg\min} \kscore({\mathbf{u_i}}) \)
    \State \# {Step 4: Compute the aggregated update}
    \State \(
        \overline{\mathbf{u}} = \frac{1}{|\mathcal{N}_{\mathbf{u}_{i^{*}}}|} \sum_{\mathbf{u}_j \in \mathcal{N}_{\mathbf{u}_{i^{*}}}} \mathbf{u}_j
        \)
    \quad \quad  \# \( \mathcal{N}_{\mathbf{u}_{i^{*}}} \) represents the set of the top 
    \( n\! -\! \hat{f}_{i^*} \) updates. 
    \State \Return \( \overline{\mathbf{u}} \)
\EndProcedure
\end{algorithmic}
\end{algorithm}

\subsection{Enhancing Stability}
We also notice that the standard \Krum algorithm (Step 3 in Section 1) selects only the single client update with the smallest \kscore as the aggregated result. However, this approach can be unstable, especially when client data are Non-IID. Relying on a single update while discarding other honest model updates can lead to a less representative aggregated update, slower convergence, and increased communication cost.

To address this issue, we are inspired by the \mKrum algorithm~\cite{blanchard2017machine}, which averages the top \(n\! -\! f\) client updates to improve the robustness of \Krum. Based on this, we propose Average-\rKrum (\ArKrum), which first identifies the client \( C_{i^*} \) with the smallest \kscore, selects the top \( n\! -\! \hat{f}_{i^*} \) client updates closest to its update \( \mathbf{u}_{i^*} \), and averages these updates to obtain a more representative aggregated update \(\overline{\mathbf{u}} = \frac{1}{|\mathcal{N}_{\mathbf{u}_{i^{*}}}|} \sum_{\mathbf{u}_j \in \mathcal{N}_{\mathbf{u}_{i^{*}}}} \mathbf{u}_j\), where \( \mathcal{N}_{\mathbf{u}_{i^{*}}} \) represents the set of the top  \( n\! -\! \hat{f}_{i^*} \) updates. By incorporating more client updates, we can improve the stability of our \rKrum algorithm. By filtering out the extreme model updates using \Alg~\ref{alg:extreme_value_filtering} and averaging the top client updates, we obtain our proposed algorithm \ArKrum, which is presented in \Alg~\ref{alg:ArKrum}. 

\section{Numerical Experiments}
\label{sec:experiments}
In this section, we evaluate the performance of our proposed \ArKrum algorithm, comparing it with other aggregation methods. We also demonstrate that \ArKrum is more stable than \rKrum (our prior work). All experiments are conducted on two public datasets and under three widely studied Byzantine attacks.

\noindent\textbf{Aggregation Algorithms:}
For a clear understanding of our numerical experiments, we reintroduce the aggregation algorithms used in this paper. 

\begin{itemize}[topsep=3pt, itemsep=3pt, parsep=0pt, partopsep=0pt]
    \item \Mean (standard aggregation): The \Mean algorithm calculates the mean of each coordinate across all client models and uses the resulting mean vector as the aggregated result.
 
    \item \Krum (robust aggregation): The \Krum algorithm first computes the pairwise squared Euclidean distances between all client updates. For each client, it then calculates the \kscore. The model update with the lowest \kscore is selected as the aggregated result.

    \item \mKrum (robust aggregation): \mKrum is an extension of \Krum. Based on the result of the standard \Krum algorithm, it identifies the client update with the lowest \kscore. Then it averages the top \( (n\!-\!f) \) client updates that are closest to this selected update. This average is used as the aggregated result.

    \item \rKrum (robust aggregation): Our prior work, \rKrum, starts by computing pairwise squared Euclidean distances between all client updates as \Krum. Then, for each client, it estimates \(f\) using the SSE-based method. After estimating \(f\), it computes \kscore and selects the update with the lowest \kscore as the aggregated result.

    \item \ArKrum (robust aggregation): \ArKrum, introduced in this paper, builds on \rKrum. It first computes pairwise squared Euclidean distances as \rKrum. For each client, it filters out extreme updates and estimates \(f\). Then it computes \kscore, selects the update with the lowest \kscore, and averages the \( (n\! -\! \hat{f}) \) updates closest to this selected update. The average is used as the aggregated result. Here \(\hat{f}\) is the estimated \(f\). 
\end{itemize}

\subsection{Experimental Setup}
In this section, we introduce the datasets, the data generation process, and the Byzantine attacks considered in our experiment.

\noindent\textbf{Datasets:} We assess the performance of each aggregation algorithm on two publicly available datasets: MNIST and SENTIMENT140.
\begin{itemize}[topsep=3pt, itemsep=3pt, parsep=0pt, partopsep=0pt]
    \item \textbf{MNIST}~\cite{lecun1998gradient}: MNIST is an image dataset designed for multiclass classification. The dataset consists of 60,000 training images and 10,000 testing images. Each 28×28 grayscale image represents a handwritten digit from 0 to 9.
 
    \item \textbf{SENTIMENT140}~\cite{go2009sentiment140}: SENTIMENT140 is a text dataset designed for sentiment analysis. The dataset contains 1,600,000 tweets, each annotated with sentiment labels: 0 for negative sentiment and 4 for positive sentiment. 
\end{itemize}

\noindent\textbf{Data Distribution:} We considered two data distributions: Independent and Identically Distributed (IID) and Non-Independent and Identically Distributed (Non-IID). 
\begin{itemize}[topsep=3pt, itemsep=3pt, parsep=0pt, partopsep=0pt]
    \item \textbf{Independent and Identically Distributed (IID):} In the IID setting, each client draws samples independently from the same global data distribution. This results in the data distribution being identical across clients. To simulate this scenario, we generate data for each client using a Dirichlet distribution with the parameter \( \alpha\!=\!10 \) ~\cite{ng2011dirichlet}. A higher value of \( \alpha \) ensures that the data distributions across clients are more similar to each other.

    \item \textbf{Non-Independent and Identically Distributed (Non-IID):} 
    In the Non-IID setting, each client also draws samples independently, but from different distributions based on the global data distribution. This results in different data distributions across clients. To simulate this, we use a Dirichlet distribution with \( \alpha\!=\!0.5 \). A lower value of \( \alpha \) results in more diverse data distributions across clients.

\end{itemize}

\noindent\textbf{Attacks:} We evaluate the robustness of our proposed \ArKrum aggregation algorithm under three widely studied Byzantine attacks. 
\begin{itemize}[topsep=3pt, itemsep=3pt, parsep=0pt, partopsep=0pt]
        \item \textit{Large Outlier}: We do not restrict the behavior of Byzantine clients. They can generate and send any arbitrary (even extreme) model updates to the central server. To simulate this scenario, we let Byzantine clients produce large abnormal updates by sampling from a Gaussian distribution with mean \( \mu\!=\!0 \) and a large standard deviation \( \sigma\!=\!10 \). 

        \item \textit{Noise Injection}: Byzantine clients inject random noise to their model updates. Here, the random noise is sampled from a standard normal distribution \( \mathcal{N}(0,1) \), where \( \mu\!=\!0 \) and a smaller \( \sigma\!=\!1 \) compared to the \textit{Large Outlier} attack. These random noises are less extreme than the \textit{Large Outlier} attack but can still pose challenges for aggregation methods.
        
        \item \textit{Label Flipping}: Byzantine clients flip their data labels, train their models on the corrupted data, and send the resulting updates to the central server. 
        
\end{itemize}

\noindent\textbf{Implementation Details:} We implement all our code in Python 3.10.14, using PyTorch 2.6.0 as the primary deep learning framework. We conduct all experiments on a high-performance computing cluster (NVIDIA DGX SuperPOD) at Southern Methodist University~\cite{smu_odonnell_institute}. Our framework simulates an FL process as in \Fig~\ref{fig:fl_architecture}, where \( n\!=\!100 \) clients independently train their local models for local datasets.
Among these clients, \(f\!=\!48 \) clients are Byzantine, and the remaining \( h\! =\! n\!-\!f\! =\! 52\) clients are honest. The choice \(f\!=\!48\) reflects the maximum number of Byzantine clients that the standard \Krum algorithm can tolerate under the constraint \(\!2\!+\!2f\!<\!n\).
We train a global model within our framework for 200 communication epochs between the central server and clients. 
During each communication epoch, each client trains its local model for five epochs. 

\subsection{Experimental Results} 

\textbf{MNIST:}
We evaluate our proposed \ArKrum algorithm on the MNIST dataset under three types of attacks. 
Since this is an image dataset, our goal is to train a CNN model to classify each image into one of the ten digit classes.
The CNN model consists of five layers, where the first three are convolutional layers, and the last two are fully connected layers. The activation function employed throughout the network is Leaky ReLU, with a negative slope of 0.2. 

%
\begin{figure}[ht]
     \vskip -0.5cm  
    \centering
    \includegraphics[width=\textwidth]{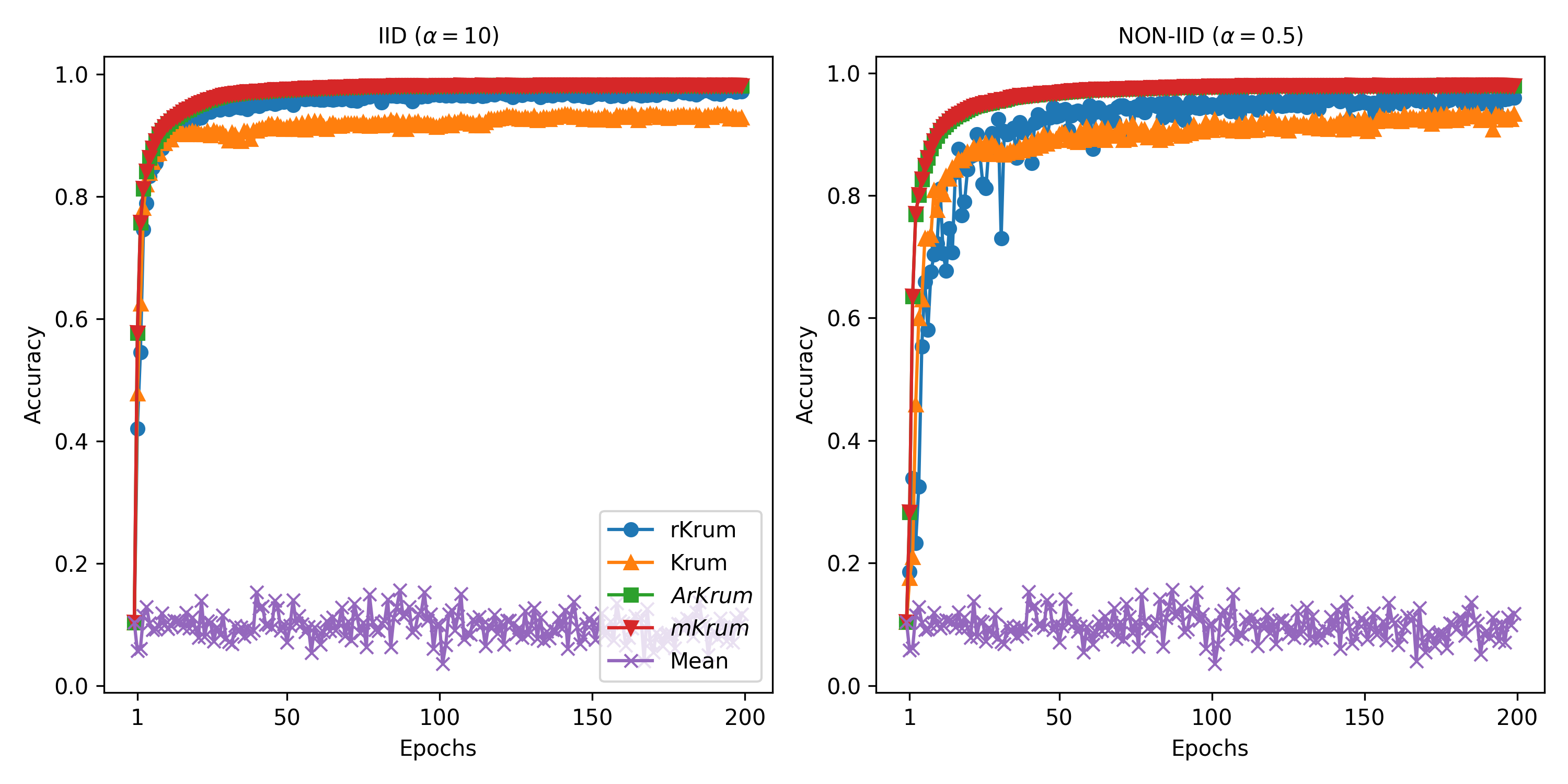}
    \caption{Performance comparison of five aggregation algorithms on IID vs. Non-IID settings under the \textit{Large Outlier} attack on the MNIST dataset.}
    \label{fig:mnist_large_outlier}
    \vskip -0.5cm  
\end{figure}

\vspace{0.1cm}
\noindent{\textit{Large Outlier}:}
Figure~\ref{fig:mnist_large_outlier} presents the classification results under the \textit{Large Outlier} attack. 
The left shows the performance of five aggregation algorithms in terms of accuracy under the IID setting (with \( \alpha\!=\!10 \)), while the right shows the performance under the Non-IID setting (with \( \alpha\!=\!0.5 \)). 
In the IID setting, all \Krum-based algorithms outperform the \Mean method in both accuracy and stability. \ArKrum and \mKrum achieve the best results, showing the highest accuracy and most consistent performance. \rKrum follows closely, while the standard \Krum algorithm performs the worst among all the \Krum-based methods.
These results demonstrate the benefit of aggregating multiple client updates (as done in \ArKrum and \mKrum) for improving the robustness and stability of \rKrum and \Krum.
In the Non-IID setting (right of \Fig~\ref{fig:mnist_large_outlier}), 
all \Krum-based algorithms still outperform the \Mean method in accuracy, despite the presence of data heterogeneity among clients. \ArKrum and \mKrum still achieve the highest accuracy and exhibit the most stable performance. In contrast, both \rKrum and \Krum suffer performance degradation. Although \rKrum becomes less stable than \Krum, it still slightly outperforms the standard \Krum algorithm in accuracy. Again, these results demonstrate the benefit of aggregating multiple updates for improving the robustness and stability of \rKrum and \Krum even under the Non-IID setting.

\begin{figure}[ht]
    \vskip -0.2cm  
    \centering
    \includegraphics[width=\textwidth]{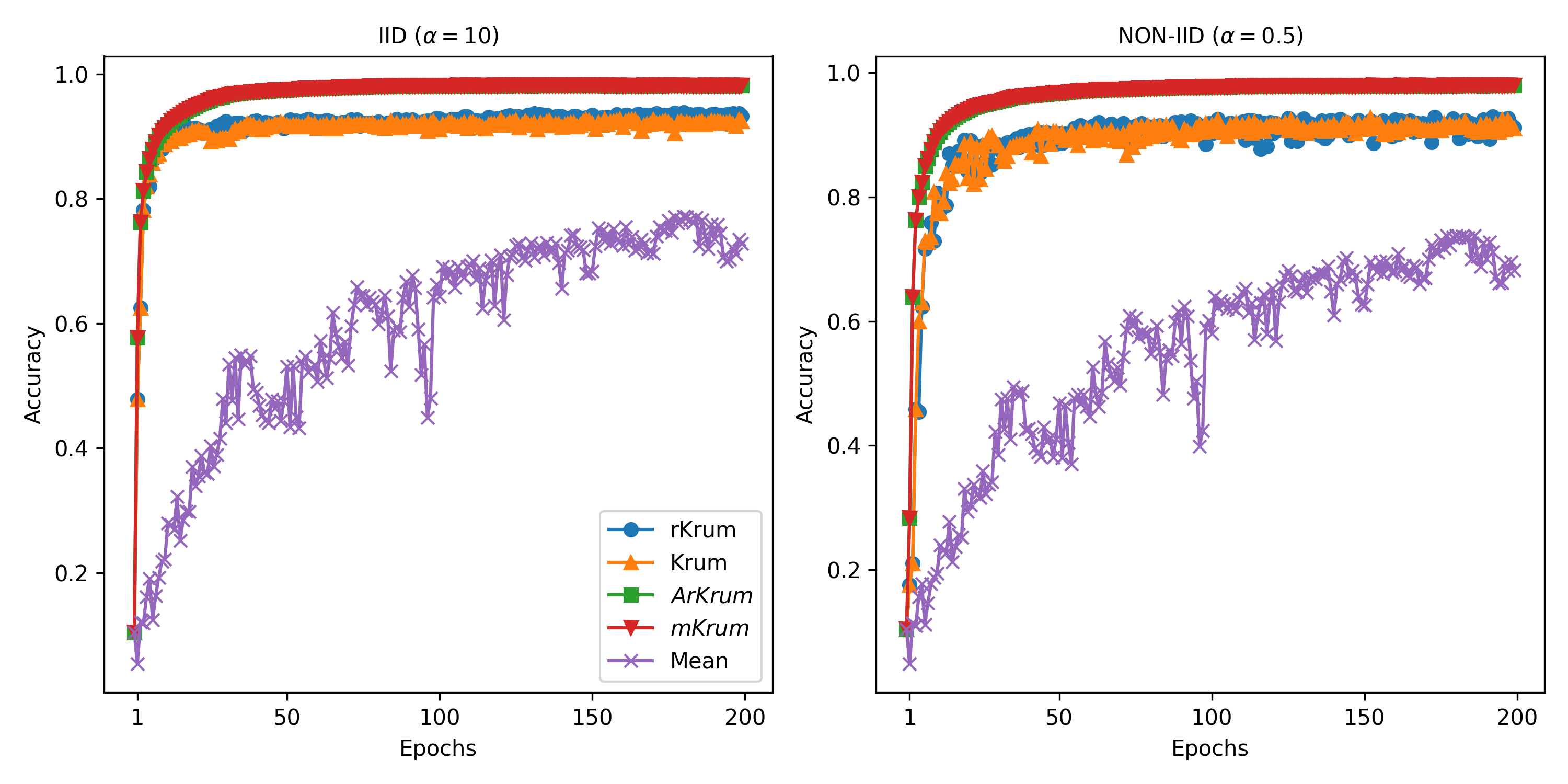}
    \caption{Performance comparison of five aggregation algorithms on IID vs. Non-IID settings under the \textit{Noise Injection} attack on the MNIST dataset.}
    \label{fig:mnist_noise_injection}
    \vskip -0.6cm  
\end{figure}
%
\vspace{0.1cm}
\noindent{\textit{Noise Injection}:}
Figure~\ref{fig:mnist_noise_injection} presents the classification results on the MNIST dataset under the \textit{Noise Injection} attack. 
%
In the IID setting (left of \Fig~\ref{fig:mnist_noise_injection}), all \Krum-based algorithms outperform the \Mean method in both accuracy and stability as before. 
Among all \Krum-based algorithms, \ArKrum and \mKrum again achieve the highest accuracy and exhibit the most stable performance, while \rKrum and \Krum perform comparatively worse. 
However, compared to the \textit{Large Outlier} attack, the \Mean algorithm performs better. The reason is that the injected random noise is less extreme, resulting in smaller deviations from the true model updates.
As a result, the \Mean algorithm does not completely fail in this scenario.
In the Non-IID setting (right of \Fig~\ref{fig:mnist_noise_injection}), 
all \Krum-based algorithms still outperform the \Mean algorithm in accuracy despite the presence of data heterogeneity. Among them, \ArKrum and \mKrum still achieve the highest accuracy and exhibit the most stable performance. The performance of \rKrum and \Krum degrades in this setting. Again, these results demonstrate the benefit of aggregating multiple updates for improving the robustness and stability of \rKrum and \Krum.

\vspace{0.1cm}
\noindent{\textit{Label Flipping}:}
For the \textit{Label Flipping} attack, we flip label 0 to 9, label 1 to 8, label 2 to 7, label 3 to 6, and label 4 to 5.
\begin{figure}[ht]
    \vskip -0.6cm  
    \centering
    \includegraphics[width=\textwidth]{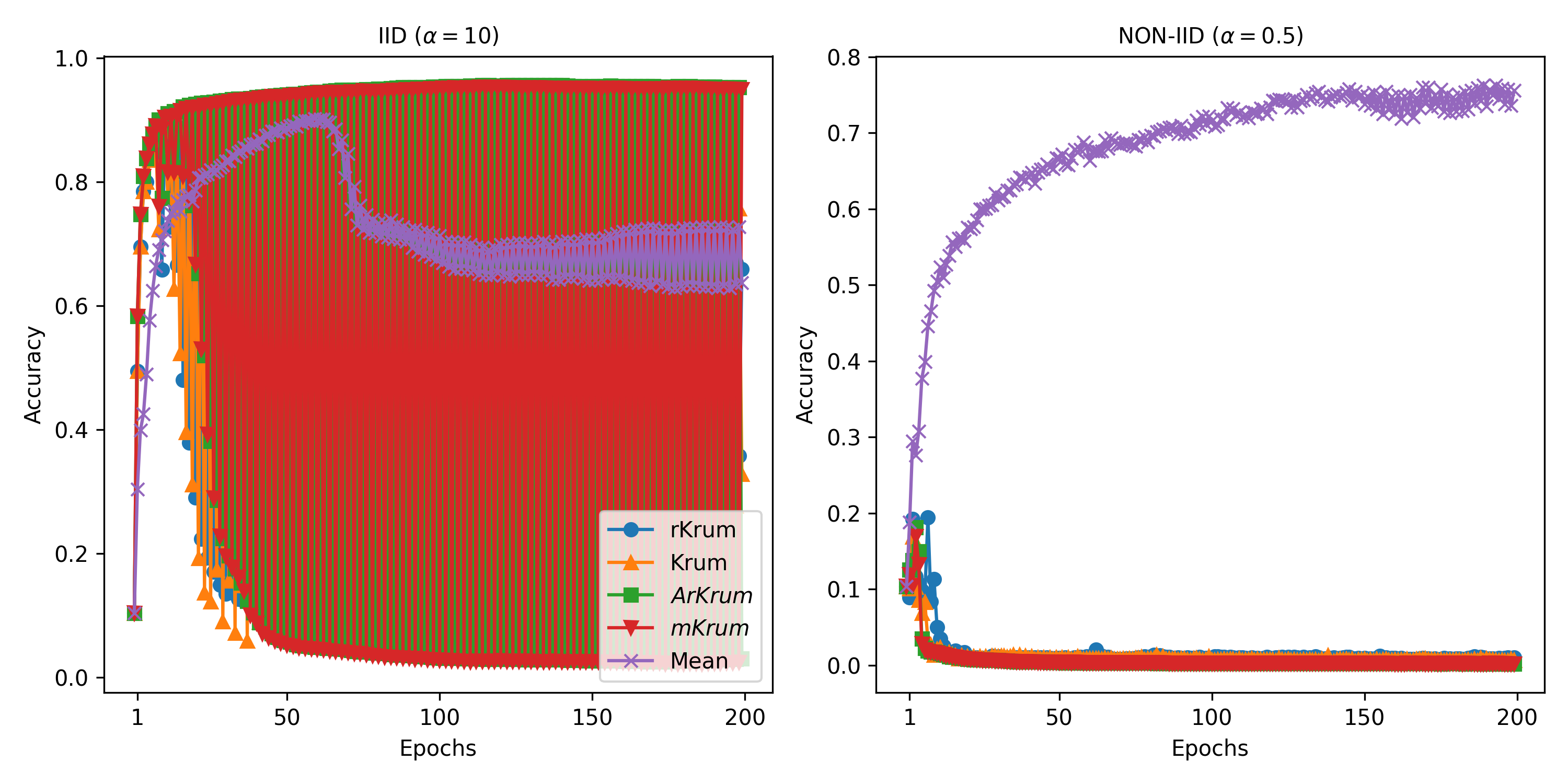}
    \caption{Performance comparison of five aggregation algorithms on IID vs. Non-IID settings under the \textit{Label Flipping} attack on the MNIST dataset.}
    \label{fig:mnist_label_flipping}
    \vskip -0.6cm  
\end{figure}
In the IID setting (left of \Fig~\ref{fig:mnist_label_flipping}), all \Krum-based methods fail due to instability caused by the misleading gradients resulting from flipped labels. Similarly, the \Mean method also fails to achieve acceptable accuracy. 
In the Non-IID setting ( right of \Fig~\ref{fig:mnist_label_flipping}), all \Krum-based methods fail again, and their performance degrades further due to data heterogeneity among clients. 
Surprisingly, the \Mean method remains effective in this case. One possible explanation is that, under Non-IID data, honest clients already generate diverse model updates. As a result, the label flipping attack does not introduce significant changes to the updates, making the effect of Byzantine updates less disruptive to the averaging process.
In contrast, \Krum and \rKrum rely on identifying updates that are close to others in Euclidean distance. Because the changes introduced by flipping labels are subtle, Byzantine updates can appear similar to honest ones and may be mistakenly selected as the aggregated result. Since \mKrum and \ArKrum build upon \Krum and \rKrum respectively, selecting a compromised update in the early stage of their aggregation leads to incorrect final updates, ultimately causing the \mKrum and \ArKrum algorithms to fail under this attack.

\vspace{0.4cm}
\noindent\textbf{SENTIMENT140:} Since this is a text dataset, we first extract the relevant features, including the tweet text and its corresponding sentiment label. Each tweet's text is then processed using the BERT (Bidirectional Encoder Representations from Transformers) model to generate fixed-length vector representations~\cite{devlin2019bert}. The text is tokenized, and each tweet is converted into a 768-dimensional vector. For the sentiment labels, we remap the original sentiment annotations (0=negative, 4=positive) to binary labels (0 for negative and 1 for positive). We randomly sample a 10\% subset of the entire dataset to improve training efficiency. The sampled data is then split into training and test sets, with 80\% of the data allocated for training and 20\% for testing. To further enhance training efficiency, we apply principal component analysis to reduce the dimensionality of the BERT embeddings from 768 to 100.
%
\begin{figure}[ht]
     \vskip -0.5cm  
    \centering
    \includegraphics[width=\textwidth]{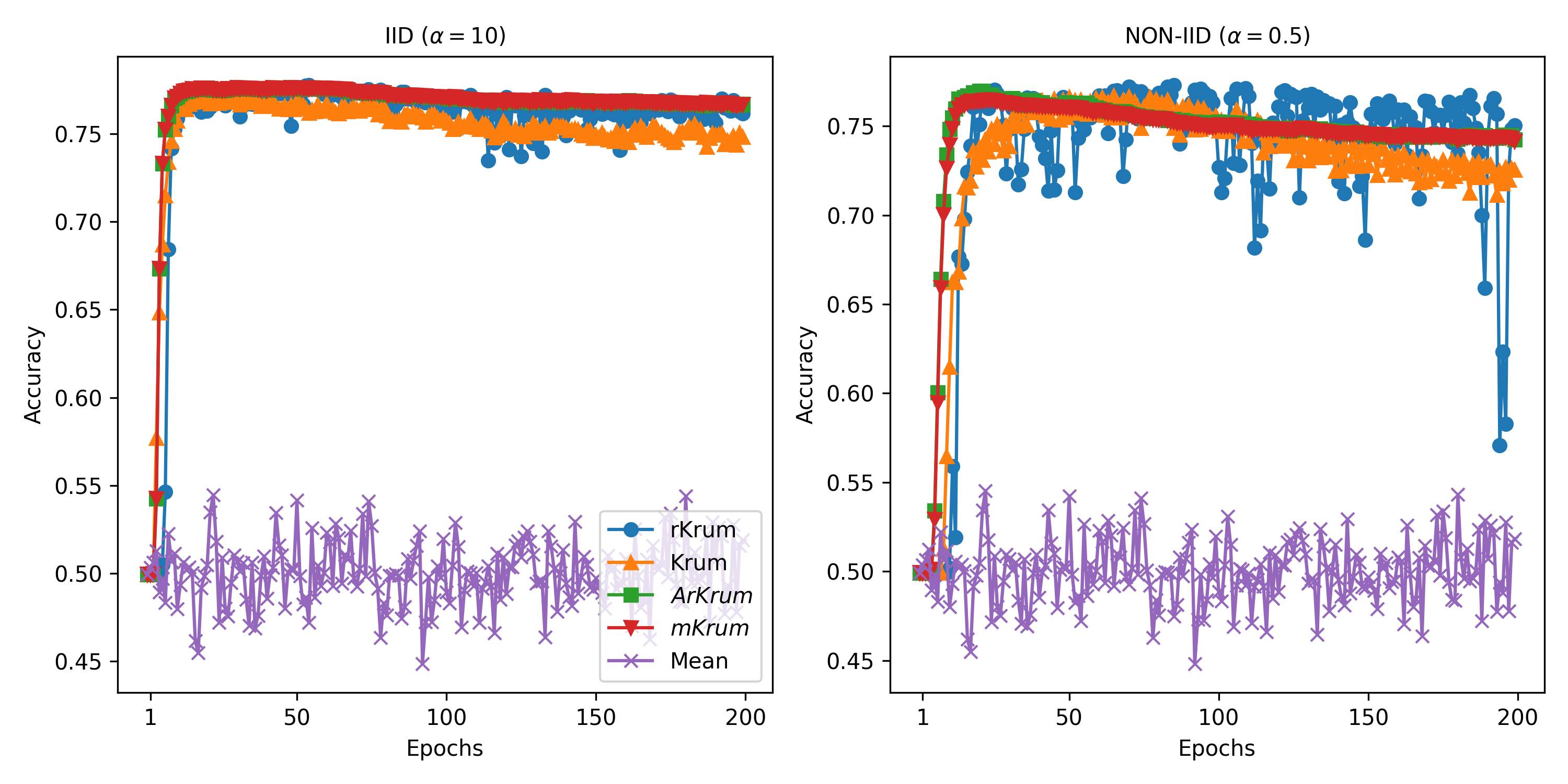}
    \caption{Performance comparison of five aggregation algorithms on IID vs. Non-IID settings under the \textit{Large Outlier} attack on the SENTIMENT140 dataset.}
    \label{fig:sentiment140_large_outlier}
    \vskip -0.5cm  
\end{figure}
%
In this experiment, our objective is to learn an FNN model to predict whether a tweet is positive or negative. The FNN model consists of four fully connected layers. 
The input layer reduces the 100-dimensional data to 32 dimensions. Then, the second and third layers further reduce it to 16 and 8 dimensions, respectively. Finally, the output layer maps the 8-dimensional features to 2 classes for prediction.

\vspace{0.1cm}
\noindent{\textit{Large Outlier}:}
In the IID setting (left of \Fig~\ref{fig:sentiment140_large_outlier}), all \Krum-based algorithms outperform the \Mean method in both accuracy and stability as before. Among them, \ArKrum and \mKrum achieve the highest accuracy and exhibit the most stable performance. While \rKrum performs slightly better than \Krum in terms of accuracy, it is less stable. Overall, both \rKrum and \Krum perform slightly worse than \ArKrum and \mKrum in both accuracy and stability. 
In the Non-IID setting (right of \Fig~\ref{fig:sentiment140_large_outlier}), 
all \Krum-based algorithms still outperform the \Mean method in accuracy, despite the presence of data heterogeneity among clients. Among them, \ArKrum and \mKrum still achieve the highest accuracy and exhibit the most stable performance. In contrast, both \rKrum and \Krum suffer performance
degradation in this setting. Again, these results demonstrate the benefit of aggregating multiple client updates for improving the robustness and stability of \rKrum and \Krum.
\begin{figure}[ht]
    \vskip -0.2cm  
    \centering
    \includegraphics[width=\textwidth]{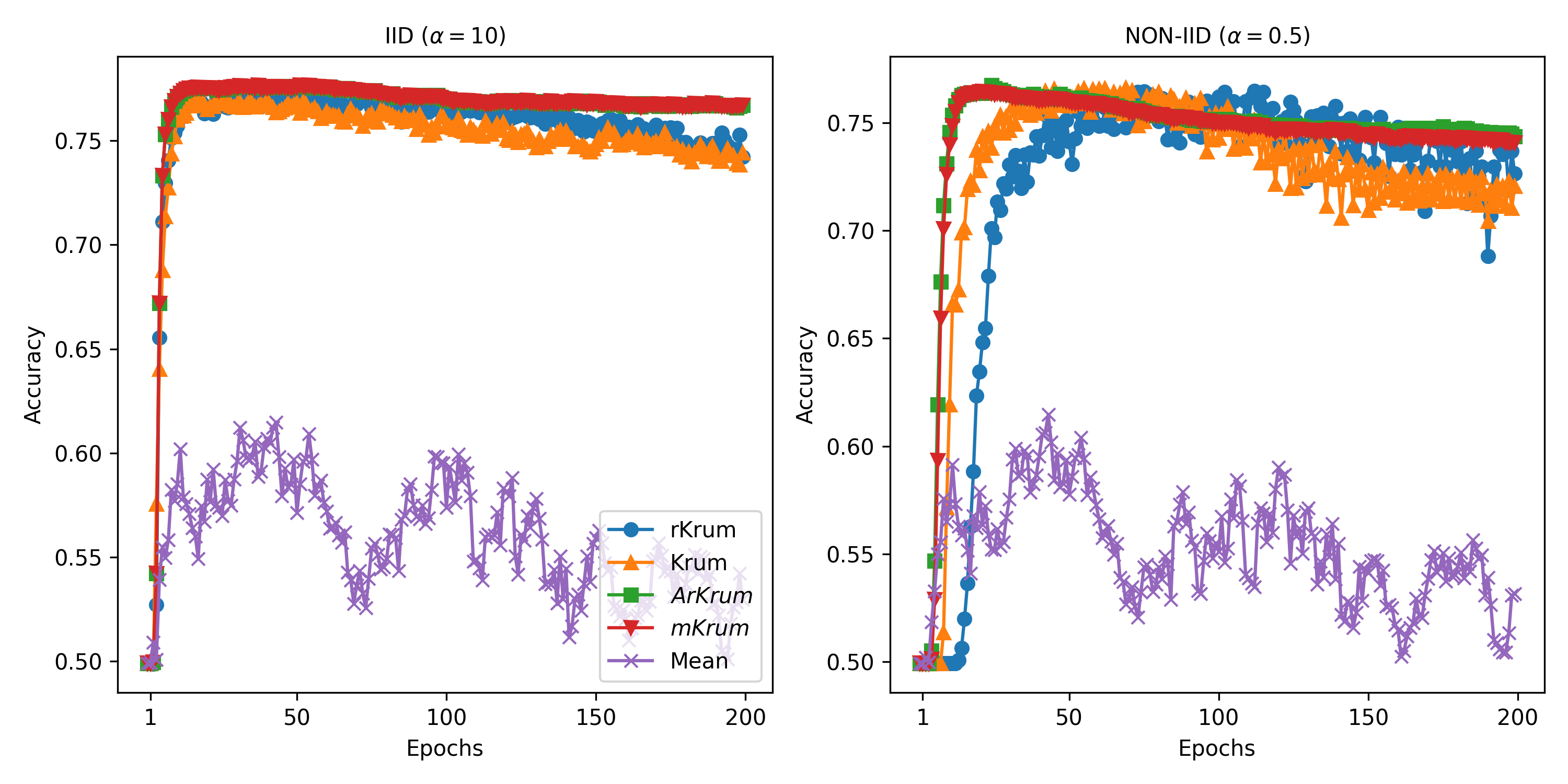}
    \caption{Performance comparison of five aggregation algorithms on IID vs. Non-IID settings under the \textit{Noise Injection} attack on the SENTIMENT140 dataset.}
    \label{fig:sentiment140_noise_injection}
    \vskip -0.6cm  
\end{figure}
%

\vspace{0.1cm}
\noindent{\textit{Noise Injection}:}
In the IID setting (left of \Fig~\ref{fig:sentiment140_noise_injection}), all \Krum-based algorithms outperform the \Mean method in both accuracy and stability as before. Among them, \ArKrum and \mKrum achieve the highest accuracy and exhibit the most stable performance. Both \rKrum and \Krum perform slightly worse than \ArKrum and \mKrum in both accuracy and stability. 
In the Non-IID setting (right of \Fig~\ref{fig:sentiment140_noise_injection}), 
all \Krum-based algorithms still outperform the \Mean method in accuracy, despite the presence of data heterogeneity among clients. Among them, \ArKrum and \mKrum still achieve the highest accuracy and exhibit the most stable performance. However, the performance of \rKrum and \Krum degrades in this setting. 
These results again demonstrate the benefit of aggregating multiple client updates to improve the robustness and stability of \rKrum and \Krum. 

\vspace{0.1cm}
\noindent{\textit{Label Flipping}:}
We flip the label 0 to 1 and the label 1 to 0. 
\begin{figure}[ht]
    \vskip -0.6cm  
    \centering
    \includegraphics[width=\textwidth]{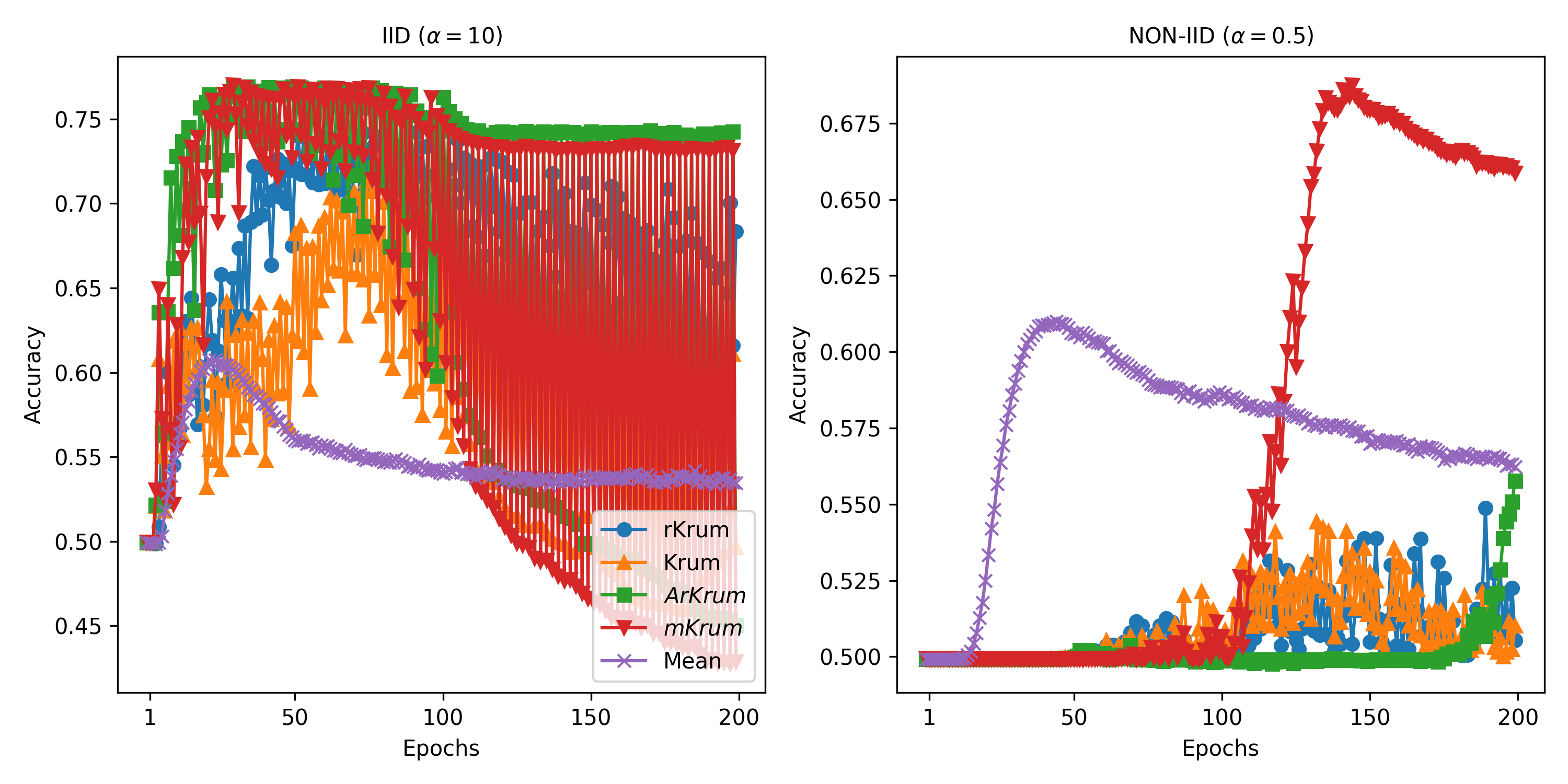}
    \caption{Performance comparison of five aggregation algorithms on IID vs. Non-IID settings under the \textit{Label Flipping} attack on the SENTIMENT140 dataset.}
    \label{fig:sentiment140_label_flipping}
    \vskip -0.6cm  
\end{figure}
In the IID setting (left of \Fig~\ref{fig:sentiment140_label_flipping}), all \Krum-based methods fail due to instability in the global model's performance. This instability is caused by the misleading flipped labels. Similarly, the \Mean method also fails to achieve an acceptable accuracy. 
In the Non-IID setting (right of \Fig~\ref{fig:sentiment140_label_flipping}), all \Krum-based methods fail again, and their performance degrades further due to data heterogeneity. The \Mean method also fails in this setting. 
All \Krum-based algorithms struggle under the \textit{Label Flipping} attack because they select the client updates that are closest (in Euclidean distance) to most others, based on the assumption that malicious updates will appear as distant outliers. However, in the \textit{Label Flipping} attack, malicious clients flip data labels (e.g., changing the label 0 to 1), causing their model updates to move in wrong but still believable directions. When multiple clients flip labels, their updates can cluster together and resemble those of honest clients. As a result, \Krum-based algorithms can mistakenly select these poisoned updates, ultimately reducing the accuracy and stability of the global model, as also observed in \cite{xhemrishi2025fedgt}.

\section{Conclusion}
In this work, we addressed a critical limitation in existing robust aggregation methods for FL, namely the dependence on prior knowledge of the number of Byzantine clients. Building on our previous \rKrum algorithm, we introduced Average-\rKrum (\ArKrum), a novel and parameter-free aggregation method that enhances both security and stability in FL systems operating under adversarial conditions. \ArKrum incorporates two key enhancements. First, it introduces a median-based filtering technique to exclude extreme model updates before estimating the number of adversarial clients. This addresses the tendency of \rKrum to underestimate Byzantine presence when extreme updates skew the estimation process. Second, \ArKrum employs a multi-update aggregation strategy, which averages multiple client updates rather than selecting a single one. This leads to more representative and stable global model updates, especially under Non-IID data distributions. We conducted comprehensive experiments on two real-world datasets (MNIST and SENTIMENT140) and evaluated performance under three representative Byzantine attack models: \textit{Large Outlier}, \textit{Noise Injection}, and \textit{Label Flipping}. Results demonstrate that \ArKrum consistently matches or exceeds the performance of existing robust methods such as \Krum, \mKrum, and \rKrum in terms of accuracy, resilience, and convergence stability. Future work will explore adaptive aggregation strategies for more complex adversarial behaviors, such as stealthy or coordinated attacks. Our findings contribute to the broader goal of making FL systems both stable and secure in real-world, adversarial environments.

\bibliographystyle{splncs04}
\bibliography{References}  

\end{document}